\documentclass[conference]{IEEEtran}
\IEEEoverridecommandlockouts
\usepackage{cite}
\usepackage{amsmath,amssymb,amsfonts}
\usepackage{graphicx}
\usepackage{textcomp}
\usepackage{xcolor}
\usepackage{array}
\usepackage{booktabs}
\usepackage{multirow}
\usepackage{url}
\usepackage[hidelinks]{hyperref}
\newcolumntype{L}[1]{>{\raggedright\arraybackslash}p{#1}}
\newcolumntype{C}[1]{>{\centering\arraybackslash}p{#1}}
\def\BibTeX{{\rm B\kern-.05em{\sc i\kern-.025em b}\kern-.08em
		T\kern-.1667em\lower.7ex\hbox{E}\kern-.125emX}}

\begin{document}
	
	\title{Policy-as-Skill: Governed LLM Decision Support with Evidence, Deterministic Control, and Audit}
	
	\author{\IEEEauthorblockN{Kabeh Mohsenzadegan}
		\IEEEauthorblockA{\textit{Institute for Smart Systems Technologies}\\
			\textit{University of Klagenfurt}\\
			Klagenfurt, Austria\\
			kabeh.mohsenzadegan@aau.at}
		\and
		\IEEEauthorblockN{Vahid Tavakkoli}
		\IEEEauthorblockA{\textit{Institute for Smart Systems Technologies}\\
			\textit{University of Klagenfurt}\\
			Klagenfurt, Austria\\
			vahid.tavakkoli@aau.at}
		\and
		\IEEEauthorblockN{Kyandoghere Kyamakya}
		\IEEEauthorblockA{\textit{University of Klagenfurt / Inst. f. Smart}\\
			\textit{Systems Technologies, Austria}\\
			\textit{\& Facult\'e Polytechnique}\\
			\textit{Universit\'e de Kinshasa}, DR-Congo\\
			kyandoghere.kyamakya@aau.at}}
	\maketitle
	
	\begin{abstract}
		Organizations increasingly use LLMs for policy, compliance, risk, and operational decision support, requiring evidence validation, review routing, version control, and auditability. We introduce Policy-as-Skill (PaS), a modular runtime that packages these functions as executable, versioned policy capabilities. Thirteen methods are evaluated with a fixed Gemma4 backend on 600 development tasks. PaS+Audit achieves 53.8\% exact accuracy, macro-F1 0.346, review F1 0.854, citation precision 1.000, policy-reference recall 0.984, and audit completeness 1.000, outperforming LLM+RAG on most governance and review metrics. Deterministic control raises aggregate accuracy to 61.2\% but is strongly task dependent, supporting selective rather than universal rule-based intervention.
	\end{abstract}
	
	
	\section{Introduction}
	Organizations increasingly use large language models for policy question answering, compliance triage, risk classification, and operational decision support. In such settings, a fluent answer is insufficient. A useful system must identify applicable policy sources, distinguish mandatory requirements from optional guidance, route uncertain or high-impact cases to accountable humans, record the policy and model versions used, and preserve enough evidence to reconstruct the decision. These requirements align with broader AI-risk and management-system guidance that emphasizes traceability, oversight, accountability, and lifecycle controls \cite{nist2023airmf,iso42001,eu2024aiact}.
	
	Retrieval-augmented generation (RAG) improves grounding by providing external documents at inference time \cite{lewis2020rag}. Policy-as-Prompt (PaP) further treats natural-language policy as a direct model instruction \cite{palla2025policyasprompt,kholkar2025policyguardrails}. However, prompt content and retrieval alone do not define how policy versions are selected, when missing evidence becomes a failure, when a human must review the case, or which fields must exist in an auditable record. Recent prompt-governance work questions whether natural-language instructions can serve as stable governance mechanisms across changing technical and institutional contexts \cite{neumann2026promptgovernance,amoore2025politicsprompt}.
	
	We therefore study \emph{Policy-as-Skill} (PaS), an architectural abstraction in which policy is packaged as a reusable, versioned, executable capability rather than only as text. The core idea is intentionally modular: skill-scoped retrieval can be used without deterministic decision override; audit/validation can be added without changing the model decision; and a controller can be enabled only where policy semantics are sufficiently explicit. This decomposition is important because a governed system can appear stronger under a composite metric simply by emitting more metadata, while a deterministic controller can improve some task types and degrade others.
	
	This paper makes five contributions. First, it defines a formal policy-skill contract and maps every component to an executable runtime artifact. Second, it provides a reproducible implementation that separates skill retrieval, deterministic control, and audit validation into clean ablations. Third, it evaluates thirteen methods over 7,800 task-method runs using a fixed model backend and primary metrics that do not reward PaS-specific metadata. Fourth, it performs controller-intervention and composite-weight sensitivity analyses, showing that the controller's aggregate gain is highly task dependent. Fifth, it explicitly distinguishes \emph{model-level unseen} data from \emph{system-level held-out} evaluation, avoiding the inaccurate claim that benchmark instances were used to train or fine-tune the LLM.
	
	The open implementation and experiment artifacts are available at \url{https://github.com/vtavakkoli/policy-as-skill}.
	
	\section{Related Work}
	RAG augments generation with non-parametric evidence and is a standard mechanism for knowledge-intensive NLP \cite{lewis2020rag}. Policy-facing applications extend this idea from generic knowledge retrieval to interpretation of normative documents. Recent studies use prompting and LLMs for privacy-policy analysis, explainability, and policy understanding \cite{goknil2024privacy,chen2025privacyanalysis,chen2024disinformation}. These works demonstrate that LLMs can extract and reason over policy text, but evidence retrieval alone does not define runtime precedence, failure behavior, review routing, or the audit record required to reconstruct a governed decision.
	
	PaP makes the governance policy itself part of the model instruction. Palla et al. study policy-as-prompt for content moderation, while Kholkar and Ahuja frame governance rules as prompt-level agent guardrails \cite{palla2025policyasprompt,kholkar2025policyguardrails}. Related scholarship examines the institutional politics of prompting, prompt governance, and collective prompting as a governance mechanism \cite{amoore2025politicsprompt,neumann2026promptgovernance,mushkani2025promptcommons}. Prompt behavior can also be sensitive to language and framing, including in ethical-reasoning settings \cite{agarwal2024ethical}. Together, this literature motivates our decision not to equate a natural-language system prompt with an enforceable policy control. PaS instead treats the prompt as one component of a versioned executable contract that also specifies evidence, validation, escalation, failure, and audit requirements.
	
	A second neighboring literature studies prompts as mechanisms for reasoning and action. Chain-of-thought prompting elicits intermediate reasoning traces \cite{wei2022cot}; ReAct interleaves reasoning and acting \cite{yao2022react}; autonomous multi-agent prompting extends these patterns to coordinated agents \cite{wang2025autonomous}; and code-as-policies work uses language-model outputs to compose robotic manipulation skills \cite{arenas2024promptrobot}. These approaches inform the decomposition of generation, action, and control, but they generally optimize task execution rather than organizational-policy accountability.
	
	Prompt optimization has likewise been formulated as a learned decision problem. Examples include reinforce-learned clarification questions \cite{yan2023askmore}, policy-gradient prompt learning \cite{lu2022dynamic}, policy-gradient discrete prompt generation \cite{li2024dialogue}, multi-objective reinforcement learning for prompt optimization \cite{jafari2024morl}, and reinforcement-learning-based automatic prompt tuning \cite{kwon2024stableprompt}. A related set of works uses the word \emph{policy} in the control or recommendation sense, including MCTS dialogue-policy planning \cite{yu2023promptmcts}, zero-shot policy learning \cite{song2024minimalist}, hierarchical prompt decision transformers \cite{wang2025hierarchical}, contrastive prompt ensembles for embodied policy adaptation \cite{kim2023contrastive}, and prompt-as-policy over knowledge graphs \cite{wang2025promptkg}. We distinguish this action-policy meaning from the governance-policy meaning in PaS: our policy object represents organizational or regulatory constraints to be retrieved, validated, audited, and selectively enforced at runtime.
	
	Finally, AI governance frameworks emphasize human oversight, traceability, accountability, risk management, and lifecycle change control \cite{nist2023airmf,iso42001,eu2024aiact}. PaS operationalizes these concerns at the application layer by binding policy scope, evidence requirements, review triggers, failure rules, and audit fields to each policy capability rather than relying on the model prompt alone.
	
	\section{Policy-as-Skill Architecture}
	We define a policy skill as
	\begin{equation}
		S=\langle n,v,R,E,D,H,A,F,P,C\rangle,
	\end{equation}
	where $n$ is skill identity, $v$ is version, $R$ is retrieval scope, $E$ is required evidence, $D$ is the decision schema, $H$ is the human-review trigger set, $A$ is the audit-field set, $F$ is failure behavior, $P$ is the structured prompt template, and $C$ is the contextual contract. The context can include data category, user role, processing location, responsible owner, or other deployment-specific fields.
	
	\begin{table}[t]
		\caption{Formal skill tuple and executable realization.}
		\label{tab:skillmapping}
		\centering
		\scriptsize
		\setlength{\tabcolsep}{2.1pt}
		\begin{tabular}{C{0.07\linewidth}L{0.29\linewidth}L{0.54\linewidth}}
			\toprule
			Term & Meaning & Runtime realization \\
			\midrule
			$n$ & skill identity & registry key and selected-skill field \\
			$v$ & version & version recorded in the governed trace \\
			$R$ & retrieval scope & allowed policy tags/sources and top-$k$ scope \\
			$E$ & evidence contract & required tags, citations, policy references \\
			$D$ & decision schema & constrained labels and schema validation \\
			$H$ & review triggers & policy-based routing/escalation rules \\
			$A$ & audit fields & evidence IDs, hashes, versions, timestamps \\
			$F$ & failure policy & unknown/review behavior on missing evidence \\
			$P$ & prompt template & structured task/evidence/output instruction \\
			$C$ & context contract & data, role, location, owner and boundaries \\
			\bottomrule
		\end{tabular}
	\end{table}
	
\begin{figure*}[t]
		\centering
		\includegraphics[width=0.96\linewidth]{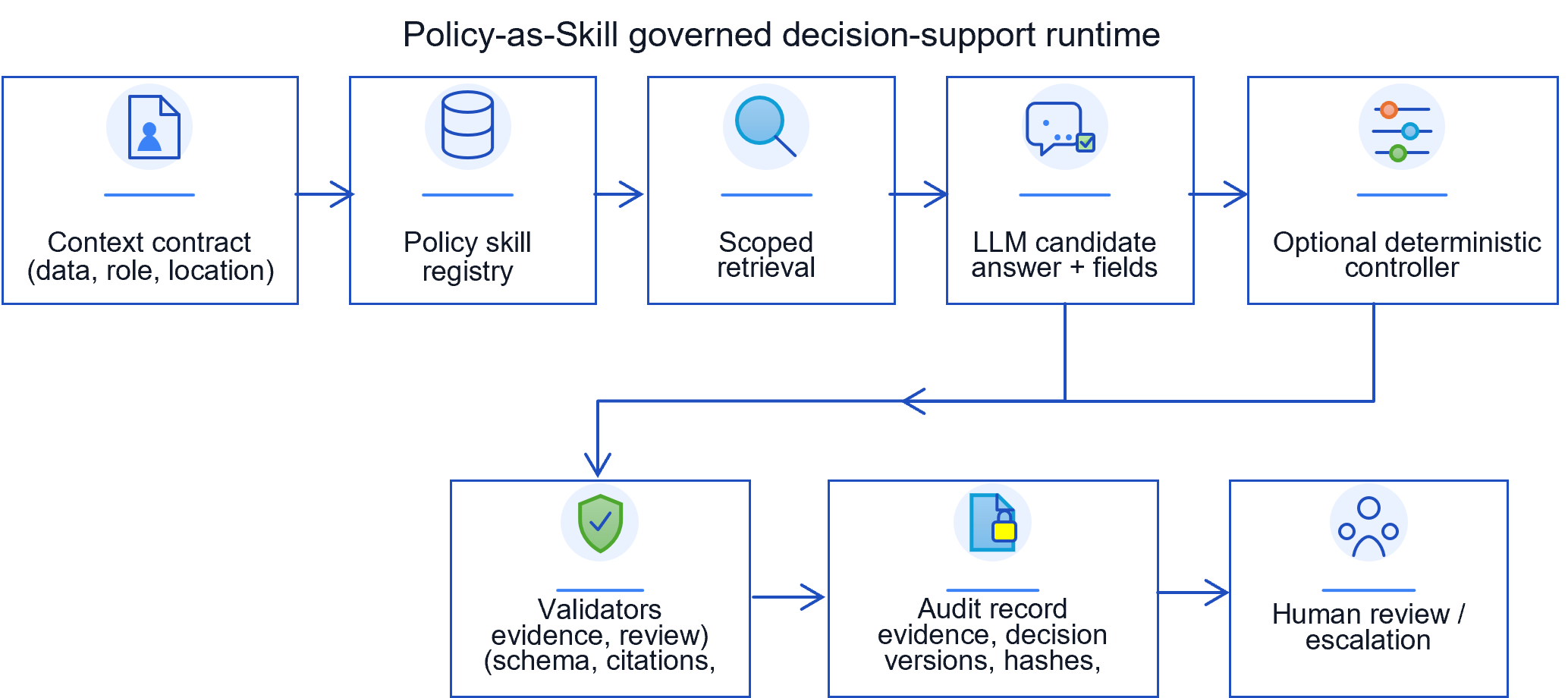}
		\caption{Policy-as-Skill runtime. Skill-scoped retrieval and audit validation are separable from the optional deterministic decision controller.}
		\label{fig:architecture}
	\end{figure*}

	The current registry implements skills for compliance checking, risk classification, conflict detection, evidence-grounded recommendation, human-review routing, and policy-version adaptation. The research runner validates that all ten tuple components are materialized in PaS traces. Importantly, the tuple does not imply that every control must override the LLM. Figure~\ref{fig:architecture} shows the modular runtime: an LLM produces a candidate answer and structured fields; validation/audit can run independently; and a deterministic controller is optional.
	
	This modularity yields four research variants: \emph{PaS Retrieval} uses skill selection and scoped retrieval; \emph{PaS+Controller} additionally applies the generic deterministic controller; \emph{PaS+Audit} applies citation/schema/audit validation without the controller; and \emph{PaS Full} combines both. The research path records that the historical benchmark-specific phrase-refinement controller is not used. The deterministic controller instead inspects retrieved normative statements and task context; it does not read task IDs, expected labels, or expected answers.
	
	\section{Evaluation Design}
	\subsection{Benchmark, Corpus, and Provenance}
	The benchmark contains 600 instances, balanced across four task families: 150 policy question-answering, 150 compliance-checking, 150 risk-classification, and 150 policy-conflict-detection cases. Reference decisions are distributed across 173 \emph{not allowed}, 155 \emph{conditional}, and 272 \emph{needs review} labels; 453 of 600 cases require human review according to the reference metadata. The conflict-detection subset is intentionally an escalation-oriented task: all 150 reference decisions are \emph{needs review}. This makes conflict detection useful for testing conflict recognition/routing but less discriminative as a multi-class decision benchmark.
	
	The policy corpus contains eleven short research fixtures spanning data protection, cloud procurement, cybersecurity/access control, workplace AI, model governance, public-sector guidance, and policy-version changes. They are synthetic, illustrative, or real-world-inspired rather than authoritative legal or enterprise policies. The EU AI Act sample is a simplified educational artifact, not the regulation itself. Reference labels are single-expert-curated and are not presented as an independently adjudicated legal gold standard.
	
	The reported run uses seed 7, 1,000 bootstrap iterations, Python 3.12.12 in Docker/WSL2, and the fixed Ollama identifier \texttt{gemma4:e2b}. Ollama was available for the run. No benchmark data were used for model parameter training or fine-tuning. At the same time, earlier trace-level analysis of the 600-task benchmark informed system/controller development. The correct provenance statement is therefore: the instances are \emph{model-level unseen} but this experiment is a \emph{system-development-informed} evaluation, not a frozen system-level held-out test. The executable benchmark, configuration, metrics, traces, and evaluation utilities are released with the research artifact.
	
	\subsection{Methods and Research Questions}
	We compare thirteen rows: Direct LLM, generic LLM, Keyword Search, Standard RAG, Hybrid RAG, Hybrid RAG with a deterministic reranker, LLM+RAG, PaP, Structured PaP, and the four PaS variants described above. All model-based methods use the same configured backend.
	
	We ask four questions. \textbf{RQ1}: How do the methods compare on exact decision correctness and human-review routing when PaS-specific audit fields receive no credit? \textbf{RQ2}: What evidence-grounding and native audit properties are added by PaS? \textbf{RQ3}: What does the controller change relative to the audit-only PaS variant? \textbf{RQ4}: Are conclusions based on the composite readiness score stable to alternative engineering weights?
	
	\subsection{Metrics and Fairness}
	Primary decision metrics are exact categorical accuracy and macro-F1 over the reference decision labels. Human-review routing is reported with precision, recall, and F1. These metrics depend only on the substantive decision/review fields and do not reward audit metadata. Citation precision measures whether emitted citations refer to retrieved evidence, and policy-reference recall measures recovery of expected policy references. The automatic evidence-faithfulness metric is retained as a diagnostic but is not treated as human validation because the manual annotation file is empty in this run.
	
	To avoid a structural advantage from richer native output schemas, the new runner also applies a \emph{common trace envelope} that every method can emit. Common trace completeness is 1.0 for all methods in the reported run. We report \emph{native audit completeness} separately as an architecture capability; unlike task accuracy, it intentionally asks whether a method natively emits the fields needed for reconstruction.
	
	The repository retains a weighted governance-readiness index for engineering diagnostics. Because its weights are design choices, it is secondary. We evaluate 286 nonnegative weight combinations on a 0.1 simplex over decision, evidence, governance, and answer-similarity components rather than selecting one weighting as a neutral correctness metric.
	
	\section{Results}
	\subsection{Primary Decision and Review Performance}
	Table~\ref{tab:mainresults} reports the primary results. PaS+Audit has the strongest balanced PaS profile without deterministic decision override: 0.538 exact accuracy (95\% bootstrap CI 0.497--0.577), macro-F1 0.346, and review F1 0.854. PaS Retrieval is similar (0.533, 0.342, and 0.855). LLM+RAG is the strongest conventional baseline by exact accuracy at 0.498 (0.458--0.537), macro-F1 0.323, and review F1 0.695. Thus, skill-scoped retrieval/audit substantially improves review routing while exact decision gains over LLM+RAG are more modest.
	
\begin{figure}[t]
		\centering
		\includegraphics[width=\linewidth]{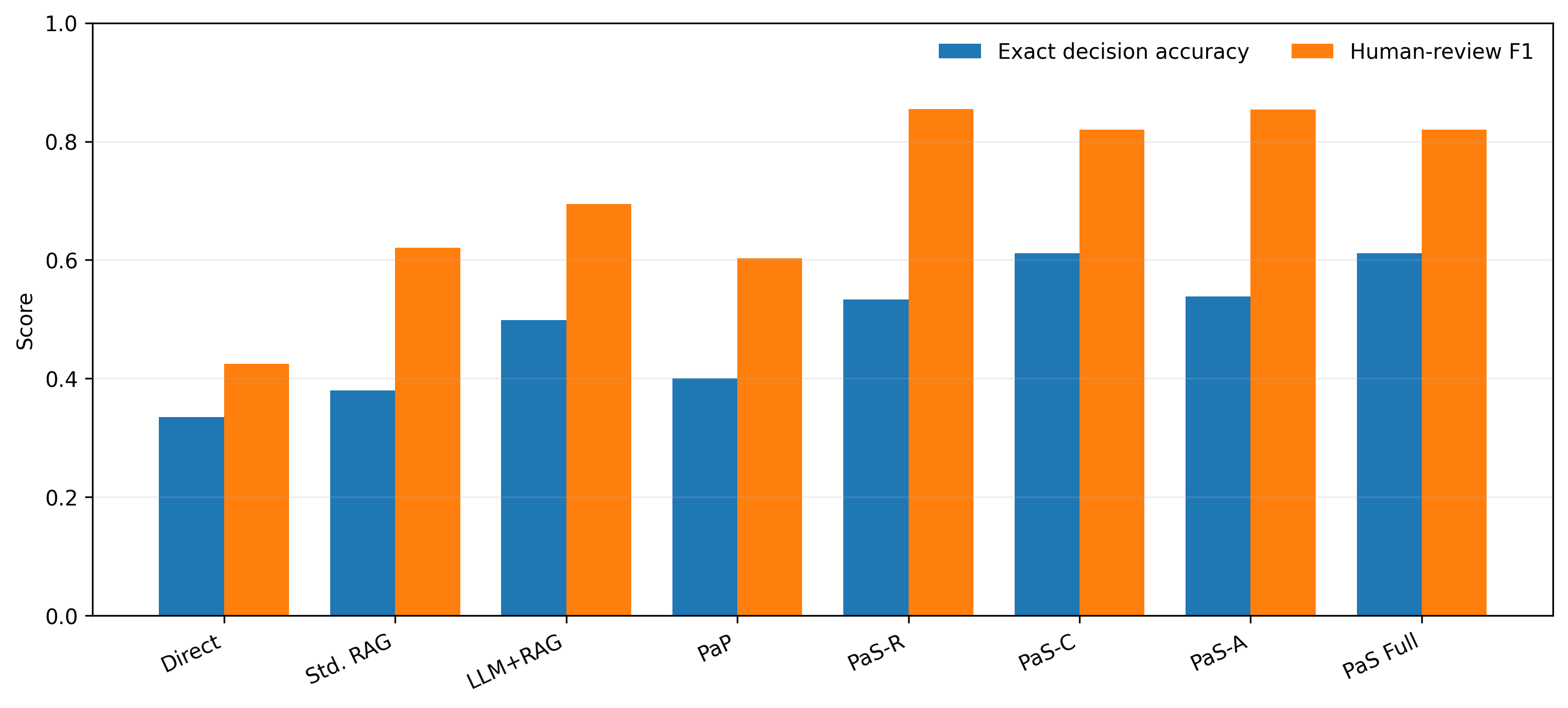}
		\caption{Exact decision accuracy and human-review F1 for representative methods. PaS audit/retrieval variants provide the strongest review routing, while the controller increases aggregate exact accuracy.}
		\label{fig:primary}
	\end{figure}

	PaS Full reaches the highest aggregate exact accuracy, 0.612 (0.573--0.652), but its macro-F1 is lower (0.306) than PaS+Audit because the deterministic controller changes the class distribution. Figure~\ref{fig:primary} visualizes this tradeoff. A paired diagnostic against LLM+RAG gives an aggregate exact-accuracy difference of +11.3 percentage points (bootstrap CI +6.3 to +16.3; exact McNemar $p=1.24\times10^{-5}$). However, this aggregate comparison must be interpreted together with the task-type analysis below because all conflict-detection reference labels are \emph{needs review}.
	
	\begin{table*}[t]
		\caption{Results on the 600-instance development benchmark. Exact accuracy, macro-F1, and review F1 are primary substantive metrics. Citation precision (Cit.P), policy-reference recall (Ref.R), and native audit completeness (Audit) are separate capabilities.}
		\label{tab:mainresults}
		\centering
		\scriptsize
		\setlength{\tabcolsep}{3.0pt}
		\begin{tabular}{lcccccc}
			\toprule
			Method & Exact & Macro-F1 & Review F1 & Cit.P & Ref.R & Audit \\
			\midrule
			Direct LLM & 0.335 & 0.239 & 0.425 & 0.000 & 0.034 & 0.625 \\
			LLM & 0.295 & 0.211 & 0.490 & 0.000 & 0.026 & 0.625 \\
			Keyword Search & 0.285 & 0.186 & 0.689 & 1.000 & 0.965 & 0.812 \\
			Standard RAG & 0.380 & 0.247 & 0.621 & 0.949 & 0.959 & 0.811 \\
			Hybrid RAG & 0.375 & 0.243 & 0.609 & 0.979 & 0.961 & 0.812 \\
			Hybrid RAG + RR & 0.367 & 0.231 & 0.647 & 0.983 & 0.965 & 0.812 \\
			LLM + RAG & 0.498 & 0.323 & 0.695 & 0.994 & 0.965 & 0.812 \\
			Policy-as-Prompt & 0.400 & 0.254 & 0.603 & 0.974 & 0.959 & 0.812 \\
			Structured PaP & 0.393 & 0.247 & 0.491 & 0.987 & 0.961 & 0.812 \\
			PaS Retrieval & 0.533 & 0.342 & \textbf{0.855} & 0.990 & \textbf{0.984} & 0.874 \\
			PaS + Controller & \textbf{0.612} & 0.306 & 0.820 & 0.991 & \textbf{0.984} & 0.875 \\
			PaS + Audit & 0.538 & \textbf{0.346} & 0.854 & \textbf{1.000} & \textbf{0.984} & \textbf{1.000} \\
			PaS Full & \textbf{0.612} & 0.306 & 0.820 & \textbf{1.000} & \textbf{0.984} & \textbf{1.000} \\
			\bottomrule
		\end{tabular}
	\end{table*}
	
	\subsection{Controller Intervention Analysis}
	The clean ablation makes it possible to separate deterministic control from audit validation. Relative to PaS+Audit, PaS Full changes the model decision on 401 of 600 tasks (66.8\%). Of those interventions, 181 convert an incorrect decision into the reference class, 137 convert a correct decision into an incorrect class, and 83 change one incorrect class to another. The net improvement is therefore 44 tasks, or 7.3 percentage points overall.
	
\begin{figure}[t]
		\centering
		\includegraphics[width=\linewidth]{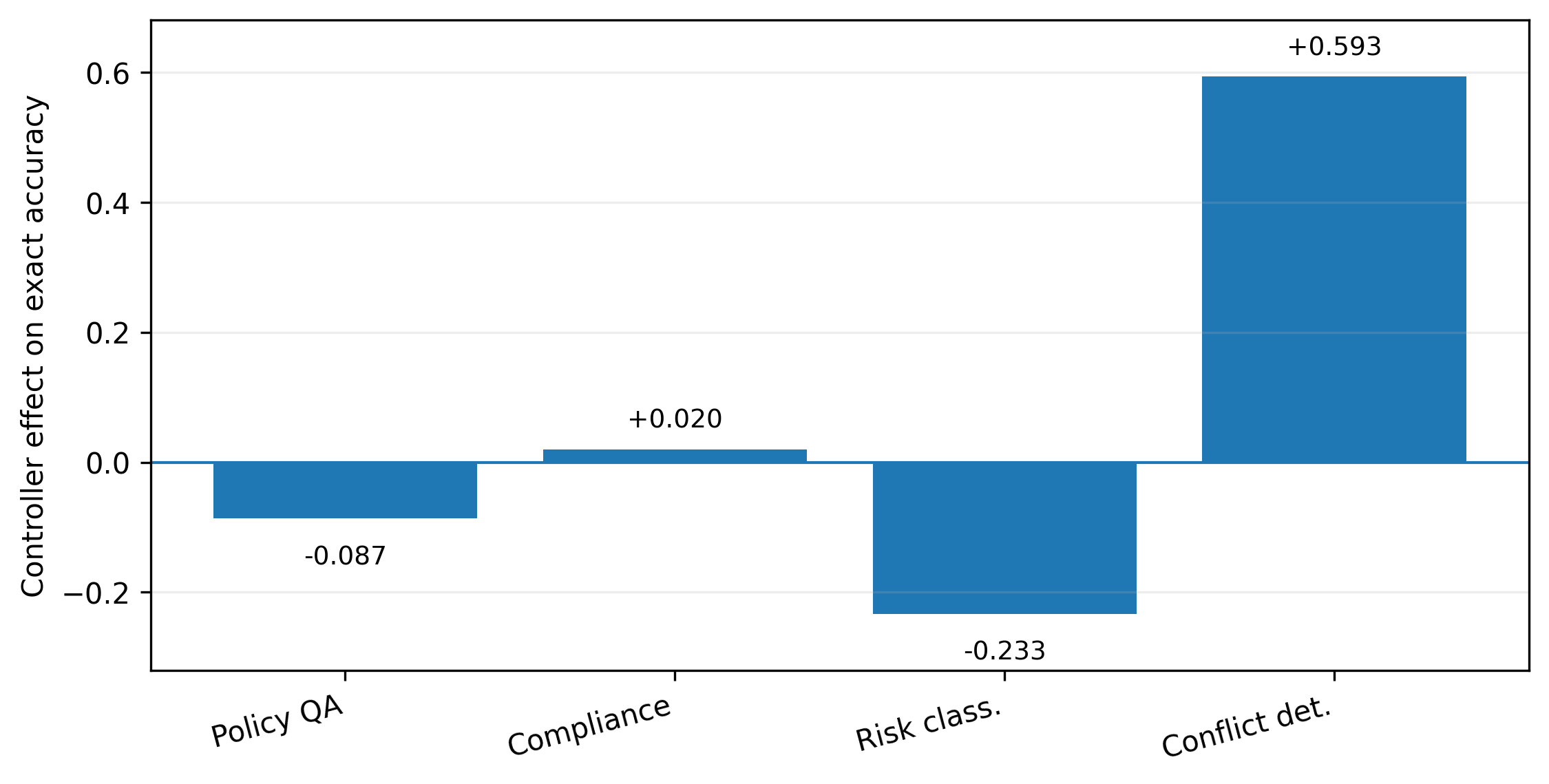}
		\caption{Task-specific effect of adding the deterministic controller to the audit-only PaS variant. The direction changes across task types.}
		\label{fig:controller}
	\end{figure}

	That aggregate gain is not uniform. Table~\ref{tab:controller} and Fig.~\ref{fig:controller} show the paired effect by task type. The controller yields +59.3 points on conflict detection because it strongly encodes conflict escalation, but it reduces risk-classification accuracy by 23.3 points and policy-QA accuracy by 8.7 points. Compliance changes only +2.0 points. The risk-classification degradation is statistically clear in this development set (paired exact test $p=1.27\times10^{-4}$), while the QA and compliance changes are not significant at 0.05.
	
	\begin{table}[t]
		\caption{Controller effect: PaS Full minus PaS+Audit exact accuracy. CIs are paired bootstrap 95\% intervals.}
		\label{tab:controller}
		\centering
		\scriptsize
		\setlength{\tabcolsep}{2.4pt}
		\begin{tabular}{lcccc}
			\toprule
			Task & Audit & Full & $\Delta$ & 95\% CI \\
			\midrule
			Policy QA & 0.407 & 0.320 & -0.087 & [-0.200, 0.027] \\
			Compliance & 0.633 & 0.653 & +0.020 & [-0.087, 0.127] \\
			Risk class. & 0.707 & 0.473 & -0.233 & [-0.347, -0.120] \\
			Conflict det. & 0.407 & 1.000 & +0.593 & [0.513, 0.673] \\
			\bottomrule
		\end{tabular}
	\end{table}
	
	A useful diagnostic is therefore to exclude the escalation-oriented conflict subset. Over the remaining 450 tasks, PaS Full achieves 48.2\% exact accuracy versus 50.4\% for LLM+RAG (difference -2.2 points, 95\% CI -7.8 to +3.3), whereas PaS+Audit reaches 58.2\% (difference +7.8 points versus LLM+RAG, CI +2.4 to +12.9). This post-hoc diagnostic reinforces the central design conclusion: the controller should be applied selectively where its policy semantics are well specified rather than treated as a universal correctness layer.
	
	\begin{figure}[t]
		\centering
		\includegraphics[width=\linewidth]{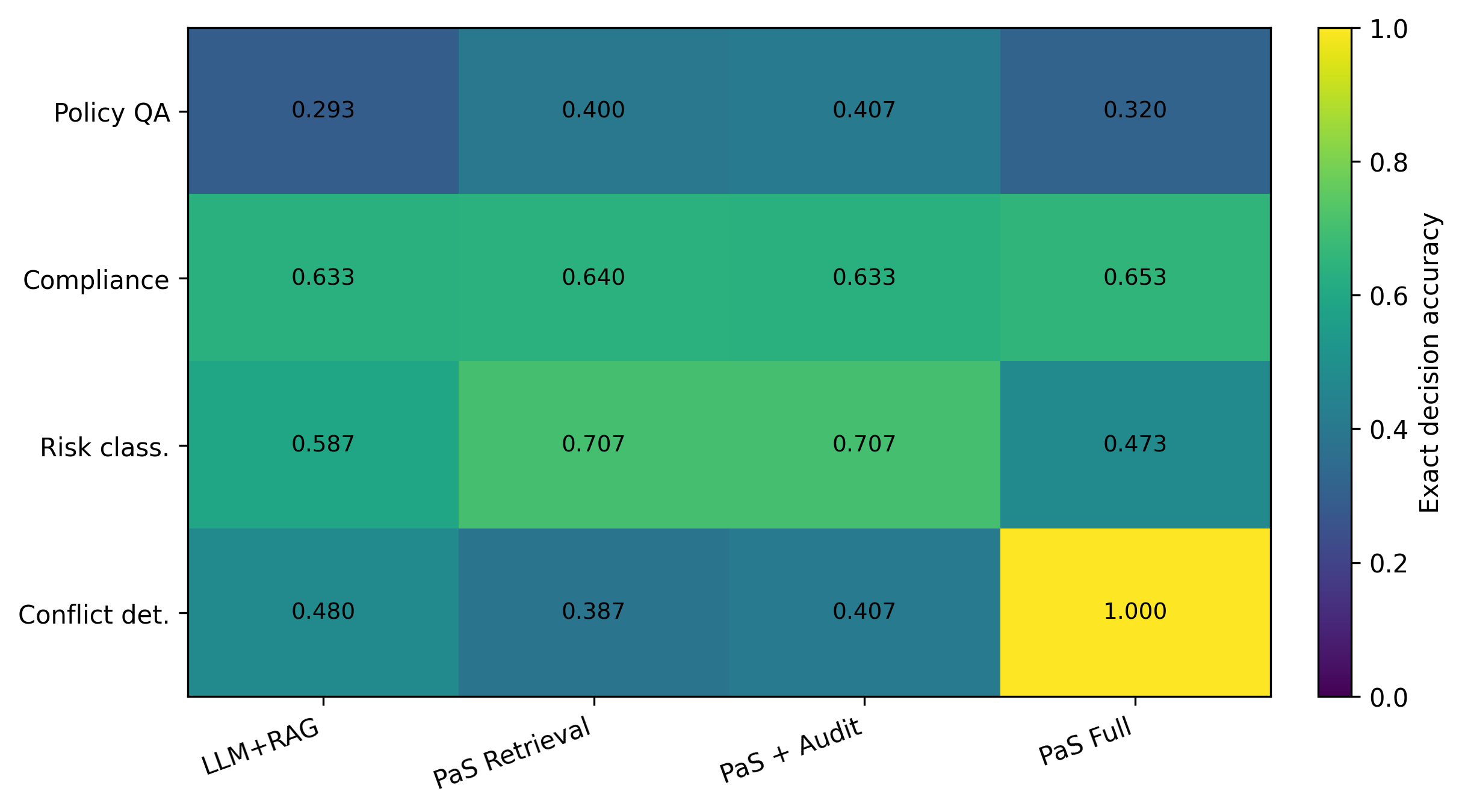}
		\caption{Exact decision accuracy by task family. The full controller is strong on the conflict-routing task but degrades risk classification relative to PaS Retrieval and PaS+Audit.}
		\label{fig:tasktype}
	\end{figure}
	
	\subsection{Evidence, Audit, and Composite Sensitivity}
	PaS+Audit and PaS Full both obtain citation precision 1.000 and policy-reference recall 0.984. LLM+RAG also has high citation precision (0.994) and recall (0.965), showing that citation validity alone does not explain the PaS review-routing gains. Keyword Search reaches extremely high automatic evidence faithfulness because it is extractive, yet its exact decision accuracy is only 0.285. This illustrates why evidence overlap and decision correctness must be reported separately.
	
	The common trace envelope is complete for all methods (1.000), so the primary comparison does not penalize baselines for missing PaS-only fields. Native audit completeness, by contrast, is an explicit architecture capability: PaS+Audit and PaS Full score 1.000, RAG/PaP variants are approximately 0.812, and closed-book LLM variants 0.625. We interpret these values as differences in reconstructability, not as decision accuracy.
	
	The secondary governance-readiness index is 0.714 for PaS Full, 0.707 for PaS+Audit, 0.681 for PaS+Controller, 0.665 for PaS Retrieval, and 0.636 for LLM+RAG. However, the 286-point weight sensitivity analysis shows that the top-ranked method changes materially with the engineering weights: PaS Full ranks first for 35.1\% of tested weightings, PaS+Audit for 30.8\%, and Keyword Search for 33.9\%; nearly all other methods are never first. This result directly demonstrates why the composite should not be the paper's primary correctness claim.
	
	Observed mean end-to-end latency is 9.75 s for PaS Full, 9.72 s for PaS+Audit, 9.86 s for PaS+Controller, and 16.65 s for LLM+RAG. These timings are descriptive only: the run used Docker/WSL2 and an external host Ollama service using CPU, not a controlled physical edge-device benchmark, and run order/caching was not designed for performance attribution.
	
	\section{Discussion}
	The new experiment changes the interpretation of Policy-as-Skill in three important ways. First, the architectural value of PaS is clearest when decision, evidence, and governance properties are kept separate. The strongest conventional baseline already retrieves and cites policy evidence well; PaS contributes a more explicit skill contract, stronger review routing, and complete native audit reconstruction. These properties are meaningful even when the underlying model decision is unchanged.
	
	Second, deterministic control is not synonymous with better reasoning. The controller is highly effective for a narrowly specified escalation task, but its broad use degrades two other task families. The result argues for \emph{policy-selective controllers}: deterministic logic should be used where a policy rule can be represented with high semantic precision (for example, explicit prohibition, missing mandatory evidence, or required escalation). Ambiguous classification and interpretive QA should preserve model uncertainty and rely more heavily on evidence, validation, and human review.
	
	Third, benchmark provenance must be described at the correct system layer. The LLM was not trained or fine-tuned on the benchmark, so calling the data ``seen by the model'' would be inaccurate. Nevertheless, system-level error analysis influenced controller development, so the 600 tasks are not a pristine held-out test of the final system. This distinction matters for governed hybrid systems in which important behavior resides outside model weights.

	\subsection{Operational Implications for Decision Support}
	From a decision-support perspective, PaS shifts governance from an unstructured prompt fragment to a lifecycle-managed capability. A policy skill binds retrieval scope, evidence requirements, decision schema, review triggers, audit fields, failure behavior, and contextual boundaries to an explicit version. This separates the model's role in interpreting evidence from the organization's role in defining decision authority. When policy changes, a new skill version can be introduced while historical traces retain the version, evidence, and routing logic that governed the original decision. This makes policy change observable and supports replay or regression analysis without silently changing the basis of earlier decisions.
	
	The results also support \emph{differentiated automation}. PaS Retrieval, PaS+Audit, and PaS Full correspond to increasingly interventionist operating profiles: evidence-grounded assistance, governed assistance with validation and reconstruction, and assistance with deterministic decision intervention. Their task-level behavior shows that no single profile should be assumed universally preferable. Explicit prohibitions, mandatory evidence requirements, and unambiguous escalation rules are suitable candidates for deterministic enforcement; interpretive QA and classification-heavy policies should preserve uncertainty and rely more on validated evidence and accountable review. In this sense, controller eligibility should be treated as a property of the policy skill rather than a global system switch.
	
	Human review is therefore a first-class governed outcome rather than an error state. Review triggers $H$ and failure behavior $F$ allow missing evidence, unresolved conflicts, or high-impact contexts to produce a visible escalation instead of forcing a model decision. The repository additionally separates development evaluation from stronger future validation through benchmark provenance, frozen-test manifests, overlap checks, human-annotation support, and physical-device measurement scripts. These mechanisms do not make the present benchmark held out, but they make subsequent validation and change management explicit and reproducible.
	
	A related management implication concerns measurement. Exact decision quality, evidence grounding, review routing, and audit completeness answer different operational questions and should not be collapsed into a single success indicator. The sensitivity analysis illustrates this directly: changing composite weights changes the preferred method. For governance reporting, organizations should therefore retain a small metric portfolio and define acceptable trade-offs per use case, rather than interpreting a single readiness score as a universal measure of correctness or deployment fitness. This also makes deployment trade-offs easier to revisit when organizational risk tolerance or review capacity changes.
	
	\section{Conclusion}
	This paper presented Policy-as-Skill as a modular architecture for governed LLM decision support. The contribution is not a claim that a policy skill makes the base LLM intrinsically more intelligent. Rather, PaS makes policy scope, evidence requirements, review triggers, failure behavior, validation, and audit reconstruction explicit and executable.
	
	The revised evaluation supports a more nuanced conclusion than a single aggregate score. Skill-scoped retrieval and audit validation provide strong review routing, citation validity, and reconstructable records without necessarily overriding the model decision. A deterministic controller can improve narrowly specified escalation behavior, but the current ablation shows that blanket controller use can reduce accuracy on risk classification and policy QA. The most defensible design is therefore modular: use deterministic controls selectively, preserve evidence and uncertainty, and keep accountable humans in the loop for ambiguous or high-impact cases. A frozen, independently validated held-out evaluation remains the next step before making system-generalization or legal-correctness claims.

\end{document}